\documentclass{article}

\usepackage[final]{neurips_2019}

\usepackage[utf8]{inputenc} 
\usepackage[T1]{fontenc}    
\usepackage{hyperref}       
\usepackage{url}            
\usepackage{booktabs}       
\usepackage{amsfonts}       
\usepackage{nicefrac}       
\usepackage{microtype}      
\usepackage{graphicx}       
\usepackage{subfig}         
\usepackage{wrapfig}        

\title{Language Representation Models for Fine-Grained Sentiment Classification}

\begin{document}

\maketitle

\begin{abstract}
    Sentiment classification is a quickly advancing field of study with applications in almost any field. While various models and datasets have shown high accuracy in the task of binary classification, the task of fine-grained sentiment classification is still an area with room for significant improvement. Analyzing the SST-5 dataset, previous work by Munikar et al. (2019) showed that the embedding tool BERT allowed a simple model to achieve state-of-the-art accuracy. Since that paper, several BERT alternatives have been published, with three primary ones being AlBERT (Lan et al., 2019), DistilBERT (Sanh et al. 2019), and RoBERTa (Liu et al. 2019). While these models report some improvement over BERT on the popular benchmarks GLUE, SQuAD, and RACE, they have not been applied to the fine-grained classification task. In this paper, we examine whether the improvements hold true when applied to a novel task, by replicating the BERT model from Munikar et al., and swapping the embedding layer to the alternative models. Over the experiments, we found that AlBERT suffers significantly more accuracy loss than reported on other tasks, DistilBERT has accuracy loss similar to their reported loss on other tasks while being the fastest model to train, and RoBERTa reaches a new state-of-the-art accuracy for prediction on the SST-5 root level (60.2\%).
\end{abstract}

\section{Introduction}

\subsection{Background}
Sentiment analysis is one of the most highly-researched natural language processing tasks with wide applications in both academia and business. The majority of sentiment-analysis research is done on the binary classification task, but fine-grained sentiment analysis allows for more precision and picks up more of the nuance in the human language. This extra granularity, however, dramatically increases the complexity involved in developing models, as sentence structure, prepositions, and dataset balance become more important. This is part of the reason why development of fine-grained sentiment analysis models using traditional sentiment-analysis tools have always lagged behind their binary counterparts.

Created by Devlin et al. at Google AI in 2019, BERT (Bidirectional Encoder Representations from Transforms), is one of the most widely used pre-training techniques for language representation and has reached state-of-the-art performances in over 11 natural language understanding tasks. BERT is powerful because it embeds words by training bidirectionally, meaning it is not restricted to reading a text left-to-right or right-to-left. This provided the model with a deeper understanding of context than its predecessors. BERT is pre-trained using two unsupervised methods: Masked LM and Next Sentence Prediction. Masked LM works to hide some of the input tokens so that they cannot “see themselves” when the model is trained in both directions. In an input sequence, 15\% of the tokens are chosen at random, 80\% of which are replaced with a [MASK] token, 10\% with a random token and 10\% with the original token. The model then predicts the values of the hidden words using cross entropy loss. BERT also trains for a next sentence prediction task in order to capture relationships between sentences. The model trains by taking in pairs of sentences A and B as input and predicting whether B immediately follows A in the original document. A and B are chosen such that B is the next sentence 50\% of the time.

In this paper we will explore proposed alternatives RoBERTa (Robustly Optimized BERT Pre-training Approach), ALBERT (A Lite BERT), and DistilBERT (Distilled BERT) and test whether they improve upon BERT in fine-grained sentiment classification.

\subsection{Alternative Language Representation Models}
\subsubsection{ALBERT}
ALBERT, which stands for “A Lite BERT”, was made available in an open source version by Google in 2019, developed by Lan et al. The model was built with the original BERT structure, but designed to drastically reduce parameters (by 89\%) without sacrificing too much accuracy. The optimized performance has been verified to produce improvements on 12 NLP tasks, including SQuAD v2.0 and RACE benchmarks. The two primary methods used to reduce the model size in ALBERT are sharing parameters across the hidden layers of the network, and factorizing the embedding layer. By having the input-level embeddings learn context-independent representations and having hidden-layer embeddings refine these into context-dependent representations, the model's capacity is more efficiently allocated.

Using the techniques described above, the ALBERT model was able to reduce the original BERT\textsubscript{BASE} model size of 108M parameters to just 12M, allowing for one to potentially scale up the size of the hidden-layer embeddings by up to 20 times.  All of this was accomplished with an accuracy reduction of 82.3\% to 80.1\% on average over the SQuAD, RACE, MNLI, and SST-2 datasets.

\subsubsection{DistilBERT}
In October 2019, Victor Sanh, Juliet Chaumond, Thomas Wolf introduced DistilBERT: a distilled version of BERT. Concerned that large models have significant environmental costs and require great computational and memory requirements, the authors propose a significantly smaller language representation model, DistilBERT, capable of similar performance to BERT in many NLP tasks with 40\% fewer parameters. The team achieves this using knowledge distillation, a compression technique in which a compact model (DistilBERT) is trained to reproduce the behavior of a larger model (BERT).

Assessing DistilBERT on the General Language Understanding Evaluation (GLUE) benchmark, the team observed that DistilBERT retained 97\% of BERT’s performance. On the Q\&A task SQuAD, DistilBERT achieved performance only 0.6\% lower than BERT in test accuracy on the IMDB benchmark.

\subsubsection{RoBERTa}
In July 2019, a joint research group between the University of Washington and Facebook AI discovered that BERT was significantly undertrained. RoBERTa, robustly optimized BERT approach, is a proposed improvement to BERT which has four main modifications. First, they trained the model longer with bigger batches, over more data. Second, they removed the next sentence prediction objective BERT has. Third, they trained on longer sentence sequences. Finally, they dynamically changed the masking pattern applied to the training data. 

Using the techniques they described, the research group tested their model on several popular language tasks. On GLUE, they reached an 88.5 average and also beat out BERT-Large on every single task. On SQuAD, RoBERTa reached 86.8 and 89.8 on the test data. Finally, for RACE, RoBERTa significantly surpassed BERT-Large by 11.2 percentage points, 83.2 vs 72.0.

\begin{table}[!ht]
  \caption{Model Comparison}
  \label{model-comparison}
  \centering
  \begin{tabular}{lrrr}
    \toprule
    \cmidrule(r){1-2}
    Model & No. Layers & No. Hidden Units & No. Self-attention Heads \cr (Total Trainable Parameters) \\
    \midrule
    BERT\textsubscript{BASE} & 12 & 768 & 12 \cr (110M) \\
    BERT\textsubscript{LARGE} & 24 & 1024 & 16 \cr (340M) \\
    ALBERT\textsubscript{BASE} & 12 & 768 & 12 \cr (12M) \\
    DistilBERT\textsubscript{BASE} & 6 & 768 & 12 \cr (66M) \\
    RoBERTa\textsubscript{BASE} & 12 & 768 & 12 \cr (125M) \\
    RoBERTa\textsubscript{LARGE} & 24 & 1024 & 16 \cr (355M) \\
    \bottomrule
  \end{tabular}
\end{table}

\section{Dataset}
The Stanford Sentiment Tree dataset was first introduced in 2013 by researchers at Stanford University who were trying to develop a dataset that would improve on standard semantic datasets by incorporating a tree structure onto the sentence to additionally capture the effects of composition on sentence semantics. The dataset includes 11855 sentences, with 215154 individual phrases. Each word, displayed individually or in 10-grams, 20-grams, and full sentences, was human-labelled on a sentiment scale of 1 to 25 (extremely negative to extremely positive). The researchers found most scores tended to center around the 5 central tick marks given, especially when considering single-words, and labellers rarely assigned extreme values, so it is generally customary to group the scores into binary classes (positive or negative) or into five classes (negative, somewhat negative, neutral, somewhat positive, positive). For given granularity n, we refer to the dataset in that form as SST-n, with SST-5 being the current standard for fine-grained sentiment classification. This dataset was developed as an alternative to classic bag-of-words datasets because it removes reliance on a few words with strong sentiment, and especially in the fine-grained cases “ignoring word order in the treatment of a semantic task is not plausible, and, as we will show, it cannot accurately classify hard examples of negation.” (Socher et al., 2013 pg 3)

\begin{wrapfigure}{r}{0.5\textwidth}
    \begin{center}
        \includegraphics[width = .48\textwidth]{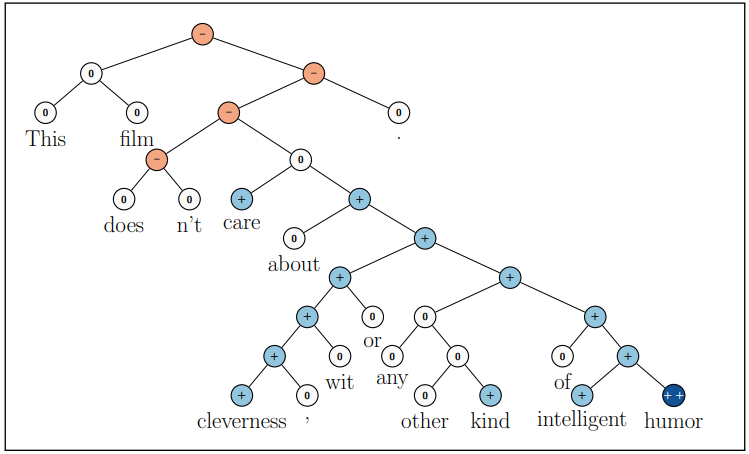}
    \end{center}
    \caption{Sample sentiment tree from SST-5}
\end{wrapfigure}

Along with creating the SST dataset, the researchers also developed a new model called a Recursive Neural Tensor Network (RNTN), which when given a vector of tokenized words represented as leaf nodes in a binary tree, would “compute parent vectors in a bottom up fashion using different types of compositionality functions.” (Socher et al. 2013, pg 4). This was the primary model used to develop the tree structures used to represent sentences, and as a result, the leaf nodes tended to join at words which were modifying phrases around it, which was a significant factor in improving the effect of negation terms in sentiment scores. At the time, the new dataset and RNTN model improved binary labelling accuracy by 5.4\% over state of the art bag-of-words models at the time (80 to 85.4\%), and improved prediction of fine-grained sentiment labels on whole phrases by 9.4\% over bag-of-features baselines (up to 80.7\%). In addition to analyzing sentences, when experimenting on targeted sentences with negation terms, the RNTN was more effective than bag of words models.

\section{Methodology}
The original paper using BERT models for sentiment classification on SST-5, ‘Fine-grained Sentiment Classification using BERT’, published by Munikar et al. in 2019, found that even without using sophisticated architecture, a model trained using BERT embeddings outperformed other popular NLP models in accuracy on both the SST-2 and SST-5 datasets at the root and total levels. In particular, compared to the original RNTN introduced by the creators of the SST dataset, they found a 3.5\% prediction accuracy increase on all nodes, and a 10\% increase on the root nodes for SST-5. The 55.5\% test accuracy on SST-5 at root level continues to be the state-of-the-art accuracy published.

For our experiment on the comparative performance of alternative BERT models on SST-5, we decided to adopt the architecture used in this paper. After pre-processing the SST text into BERT formatting which, we send the text through the pre-trained BERT embedding layer, then apply a dropout layer with probability 0.1, then finally send to a fully connected softmax layer which outputs the probability vectors for labels \{0,1,2,3,4\}. Between experiments for the 4 different BERT-like models, including original BERT, we would only change the pre-trained embedding layer to whatever model we were testing, leaving the same dropout and softmax layer. With each experiment, we ran for 30 epochs, using an ADAM optimizer with a learning rate of 1e-5 and beta values 0.9 and 0.99. After training the 4 models, using the largest-base pre-trained version available, we compare the test accuracy, test loss, and training time of the alternative BERT models against our replication of the original paper, as well as the published results of the original paper. In particular, we want to analyze the various trade-offs between accuracy and time for the BERT alternatives on a novel dataset for the alternative models.

After noticing the test accuracy tended to fluctuate randomly over 30 epochs, neither improving nor getting worse as training loss converged, and getting different patterns on different machines, we decided that the original model training process led to significant overfitting on the train set, so we decided to implement an early-stoppage protocol in our analysis. This practice also let us analyze how quickly models would converge in terms of epochs.


\section{Results}
\begin{table}[!ht]
  \caption{SST-5 Results}
  \label{results}
  \centering
    \begin{tabular}{lrrrr}
    \toprule
    \cmidrule(r){1-2}
    Model & Training Time (per epoch) & Best Test Acc. \\
    \midrule
    BERT\textsubscript{BASE} & 5:38 & 0.549     \\
    BERT\textsubscript{LARGE} & 12:38 & 0.562      \\
    ALBERT\textsubscript{BASE} & 3:16 & 0.490 \\
    DistilBERT\textsubscript{BASE} & 2:54 & 0.532  \\
    RoBERTa\textsubscript{LARGE} & N/A & 0.602     \\
    \bottomrule
  \end{tabular}
  \caption{Experiment results for classification task on SST-5 root nodes}
\end{table}
*Training Time (per epoch) is listed for training with batch size of 8 on an NVIDIA GeForce GTX 1000. RoBERTa\textsubscript{LARGE} training time not listed as training was memory intensive and required the use of a GPU from Google Colab.
\begin{figure}[!ht]
    \centering
    \subfloat[Train Loss]{\includegraphics[width = 2.5in]{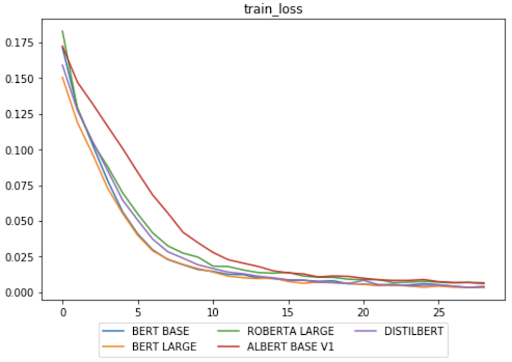}}
    \subfloat[Test Loss]{\includegraphics[width = 2.5in]{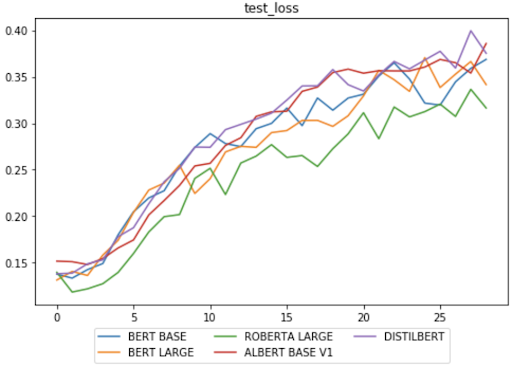}} 
    \\
    \subfloat[Train Accuracy]{\includegraphics[width = 2.5in]{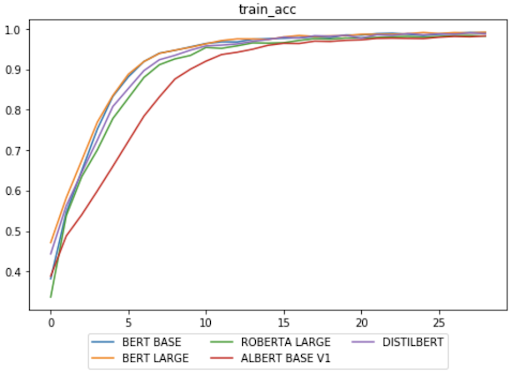}}
    \subfloat[Test Accuracy]{\includegraphics[width = 2.5in]{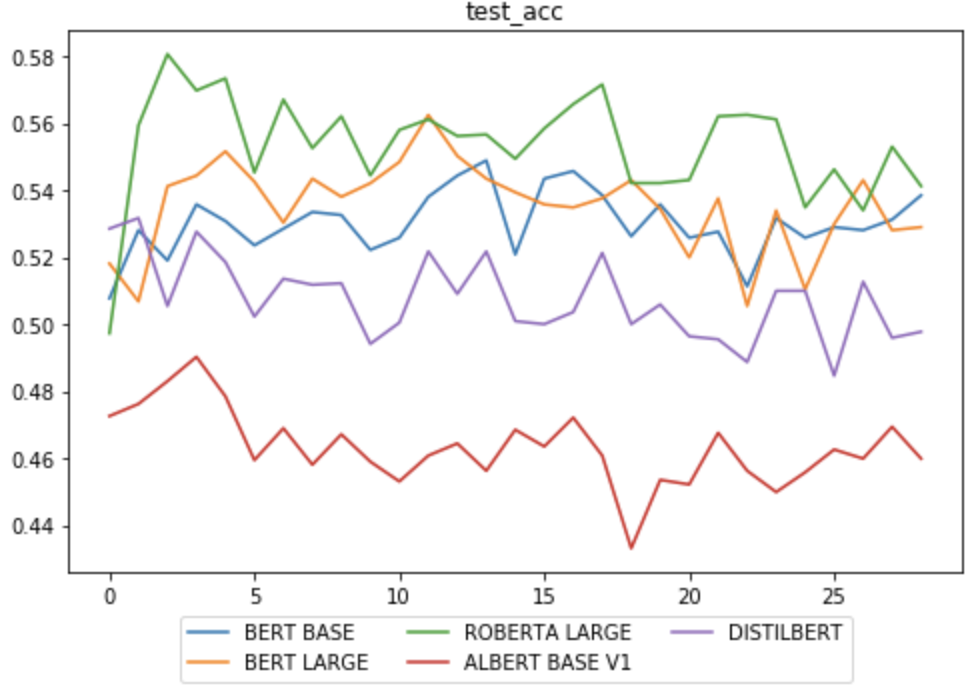}}
    \caption{Loss and accuracy per epoch plots for the 5 comparison models}
\end{figure}

\subsection{BERT}
In our replication of the BERT\textsubscript{BASE} model, the test accuracy achieved after 30 epochs was 0.538, which did not stray far from the reported accuracy of 0.532 from the original paper. However, our replication of the BERT\textsubscript{LARGE} model achieved a test accuracy of 0.529 after 30 epochs, worse than the reported state-of-the-art performance of 0.555. Upon investigating the performance of both models across epochs, we observed that after a few epochs of training, the test loss began to increase consistently. This suggested that the BERT models trained across 30 epochs on a learning rate of 1e-5 were significantly overfitting on the training data. To best compare the ability of the different language representation models, we implemented early stopping in our experiments to prevent the problem of overfitting in the original paper. The results of our implementations with early stopping are discussed below.

\subsubsection{BERT\textsubscript{BASE}}
BERT\textsubscript{BASE} achieved a 0.549 accuracy on the SST-5 test set. This was not only a significant improvement over the reported accuracy of 0.532, but was also the second-highest accuracy of all the “BASE” models.

The training for BERT\textsubscript{BASE} was slower than the other "BASE" models -- each epoch took 5 minutes and 44 seconds, and the performance did not peak until the 13th epoch whereas other models reached peak performance in 6 or fewer epochs.

\subsubsection{BERT\textsubscript{LARGE}}

As expected, BERT\textsubscript{LARGE} outperformed BERT\textsubscript{BASE} since it is larger and more computationally intensive. However, its test accuracy of 0.562 did not top RoBERTa\textsubscript{LARGE} or RoBERTa\textsubscript{BASE}. Its size came at a clear cost -- as each epoch took 12:38 to train, by far the longest of all the models.

The confusion matrix shows that BERT\textsubscript{LARGE} performed worst at the extremes. Strongly negative and and strongly positive sentiments were miscategorized as weakly negative and weakly positive sentiments, respectively, roughly half the time. Neutral sentiments were miscategorized as weakly negative roughly half the time as well.

\begin{figure}[!ht]
    \centering
    \includegraphics[width = 0.75\textwidth]{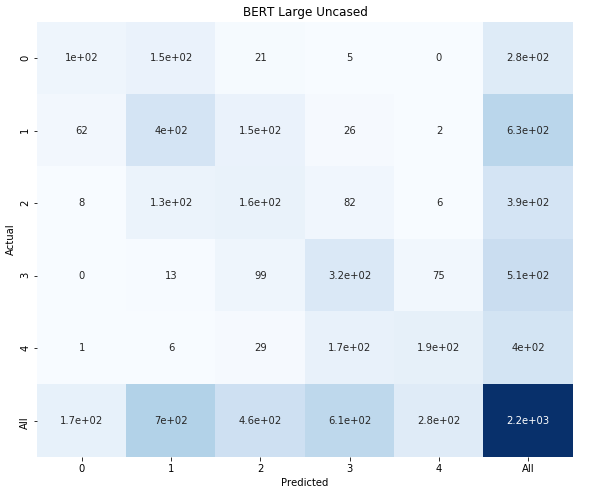}
    \caption{Confusion matrix for BERT\textsubscript{LARGE} on SST-5 test set root nodes}
\end{figure}

Additionally, the matrix shows that most of the mistakes are on the margins, meaning that a 0.562 “accuracy” understates how well the model is performing. The tendencies of miscategorization mirror the distribution of the data. Strong negatives and strong positives are underrepresented, and weak negatives are overrepresented (which is why so many neutrals are miscategorized as weak negatives.) A larger, more balanced dataset would likely yield better results.

\subsection{AlBERT}

ALBERT performed far worse than expected on the SST-5 dataset, only achieving 0.490 test accuracy. Compared to BERT\textsubscript{BASE}, the drop in accuracy from .549 to .490 is significantly greater than the 2.2\% drop found on the binary task. ALBERT did show an almost 2 minute decrease in training time per-epoch, while converging fairly quickly, but this reduction in accuracy likely outweighs the benefits of the decrease in training time, and therefore the utility of ALBERT on novel fine-grained NLP tasks is in question. The poor performance of the ALBERT model on the fine-grained task exposes many of the weaknesses the model faces and potential pitfalls as an embedding tool when generalizing to new tasks.

\subsection{DistilBERT}
Our highest performing DistilBERT model achieved a test accuracy of 0.532 whereas our replication of BERT\textsubscript{BASE} had a test accuracy high of 0.549. On the task of SST-5, the claims from the DistilBERT proposal hold up as DistilBERT retain 97\% of the BERT\textsubscript{BASE} model performance with significantly fewer parameters. Another interesting note is that the DistilBERT model training required half of the time for the BERT\textsubscript{BASE} model. In fact, the DistilBERT model required only 2:54 per epoch and was the fastest to train of all models tested. This result shows DistilBERT's distilled parameters still are effective when applied to a new task, and should be a preferred choice when requiring additional computational efficiency over BERT\textsubscript{BASE}.

\subsection{RoBERTa\textsubscript{LARGE}}

RoBERTa\textsubscript{LARGE} supplied us with our best result of 0.602 accuracy. From the original claims of the RoBERTa study, we observed that many of their claims did hold true. In particular, our RoBERT\textsubscript{LARGE} model performed exceedingly well when we trained the model using larger batch sizes. However, this comes with obvious trade-offs with computation time and memory requirements. RoBERTa\textsubscript{LARGE} was not able to train with our current hardware of a NVIDIA GeForce GTX 1080, a single graphics card was not able to handle the large number of parameters. To remedy this, we utilized cloud computing resources to try the RoBERTa\textsubscript{LARGE} model with different batch sizes, where we observed the performance increase due to larger batches.

\begin{figure}[!ht]
    \centering
    \includegraphics[width=0.75\textwidth]{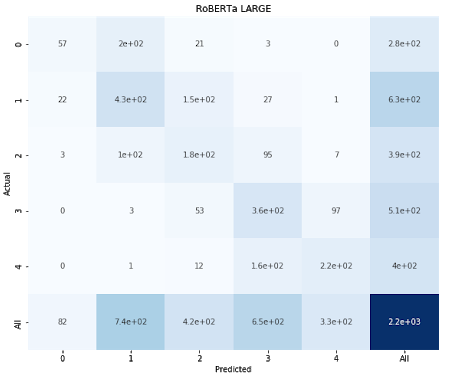}
    \caption{Confusion matrices for RoBERTa\textsubscript{LARGE} on SST-5 test set root nodes}
\end{figure}

When comparing the confusion matrices of RoBERTa\textsubscript{LARGE} and BERT\textsubscript{LARGE} models we see that performs slightly worse in the ‘Strongly Negative’ case, where the model confuses strongly and weakly negative labels. However, RoBERTa\textsubscript{LARGE} outperforms in every other case greatly, classifying 4.7\% more correct ‘Weakly Negative’ cases, 5.1\% more ‘Neutral’ cases, 7.8\% more ‘Weakly Positive’ cases, and finally 7.5\% more ‘Strongly Negative’ cases.

\section{Discussion}

When comparing performance of the alternative BERT models, the first striking observation is the significantly reduced training time of the ALBERT\textsubscript{BASE} and DistilBERT\textsubscript{BASE} models. These models have significantly less parameters (12M and 66M respectively), so as expected these training speeds come with tradeoffs in model complexity. However while ALBERT and DistilBERT perform similarly on binary-classification benchmarks, DistilBERT shows a substantially better accuracy (.532 vs .490) when applied to the fine-grained sentiment task on top of being slightly faster. 

This result likely comes down to the methods the models use to pare down parameters. ALBERT shares parameters across layers, and only trains certain layers context-dependently further reducing complexity, so while this may drastically reduce the number of parameters without sacrificing predicting power on a binary case, when increasing the output layer to five classes the rigidity shows. Meanwhile, using knowledge compression DistilBERT is trained to learn similarly to a larger model, and while it has six times the parameters of ALBERT the training time is not sacrificed. Thus, while ALBERT does significantly reduce training time and complexity if memory is an issue, there is a large sacrifice of accuracy, and DistilBERT is preferred if it is computationally feasible.

We also found that RoBERTa\textsubscript{LARGE} achieved a new state-of-the-art accuracy of .602 on the SST-5 root level after implementing early stopping. This followed the claims made in the original paper by Liu et al. (2019) that RoBERTa's increased training gave a significant increase in model accuracy over BERT\textsubscript{LARGE}. While we didn't see as dramatic of an accuracy increase as some benchmarks over BERT\textsubscript{LARGE} (.602 vs .562), RoBERTa embeddings clearly hold some performance stability when applied to the fine-grained classification task. However, there is a dramatic increase in complexity and computational intensity in training, and we found ourselves running into memory issues on the same devices that could handle BERT\textsubscript{LARGE} fine. Applied to more complex data or a higher granularity classification, the computational power may prohibit RoBERTa from being as effective. However, for this problem it clearly outperforms its competitors.

The test loss and accuracy curves shown in Figure 2 are immediate signs that the current model is overfitting on the training set. In fact, the test loss stops decreasing within 2-3 epochs. Even after incorporating early stoppage, there is a clear need for a model change to limit overfitting. Since we are using very simple model with a pre-trained embedding layer, a dropout, then a fully connected layer with a softmax activation, there are likely a few experiments that may improve performance without drastically increasing the complexity. The simplest of these would be to adjust the dropout rate, either increasing it or having it progressively increase as training continued. Similarly, we could implement a learning rate decay to prevent the model from overlearning on elements of the training sentences. 

\section{Conclusion}
From the experiments, we observe that though ALBERT and DistilBERT do not outperform BERT on fine-grained sentiment analysis, they are great alternatives for BERT\textsubscript{BASE} when memory and speed are factors of concern. In particular, the DistilBERT model was able to retain 97\% of the BERT\textsubscript{BASE} performance while only requiring half the training time. 

During replication of state-of-the-art models, the accuracy and loss plots of our experiments also revealed that the BERT models developed by Munikar et al. (2019) were overtrained and overfitting on the training set. Instead of training for 30 epochs, we found that models performed better with significantly less training, often in under 6 epochs.

By redesigning the training methodology to incorporate early stopping, we were able to achieve better performance for BERT models than the observed replication results. Furthermore, by combining early stopping with the replacement of BERT with the more optimized language representation model RoBERTa, we were successful in achieving a new state-of-the-art test accuracy of 0.602 on SST-5, overtaking the previous high of 0.555 from Munikar et al. (2019). 

\section*{References}

\small

[1] Devlin, J., Chang, M.W., Lee, K.\ \& Toutanova, K.\ (2018) BERT: Pre-training of Deep Bi-Directional Transformers for Language Understanding. arXiv:1810.04805[cs.CL]

[2] Lan, Z., Chen, M., Goodman, S., Gimpel, K., Sharma, P.\ \& Soricut., R.\ (2019) ALBERT: A Lite BERT for Self-supervised Learning of Language Representations. arXiv:1909.11942[cs.CL]

[3] Liu, Y., Ott, M., Goyal, N., Du, J., Mandar, J., Chen, D., Levy, O., Lewis, M., Zettlemoyer, L\ \& Stoyanov, V.\ (2019) RoBERTa: A Robustly Optimized BERT Pre-Training Approach. arXiv:1907.11692[cs.CL]

[4] Munikar M., Shakya S.\ \& Shrestha A.\ (2019) Fine-grained Sentiment Classification using BERT. arXiv:1910.03474[cs.CL]

[5] Sanh, V., Debut, L., Chaumond, J.\ \& Wolf, T.\ (2019) DistilBERT, a distilled version of BERT: smaller, faster, cheaper, and lighter. arXiv:1910.01108[cs.CL]

[6] Socher, R., Perelygin, A., Wu, J.Y., Chuang, J., Manning, C.D., Ng, A.Y\ \& Potts, C.\ (2013). Recursive deep models for semantic compositionality over a sentiment treebank. EMNLP, vol. 1631. pp. 1631-1642. 

\end{document}